\documentclass[final,1p,times]{elsarticle}

\usepackage{booktabs}
\usepackage{hyperref}

\makeatletter
\def\ps@pprintTitle{%
 \let\@oddhead\@empty
 \let\@evenhead\@empty
 \def\@oddfoot{}%
 \let\@evenfoot\@oddfoot}
\makeatother

\usepackage{graphicx}
\usepackage{amssymb}
\usepackage{commath}
\usepackage{mathtools}
\usepackage{subcaption}
\usepackage{float}

\DeclareMathOperator*{\argmin}{arg\,min}
\DeclareMathOperator*{\E}{\mathbb{E}}

\begin{document}

\begin{frontmatter}

\vspace*{\fill}
\begin{center}
This is a draft version, content may be revisited in subsequent versions.
\end{center}
\vspace*{\fill}
    
\title{DiffPrune: Neural Network Pruning with Deterministic Approximate Binary Gates and $L_0$ Regularization}
\author{Yaniv Shulman}
\address{yaniv@aleph-zero.info}

\begin{abstract}
Modern neural network architectures typically have many millions of parameters and can be pruned significantly without substantial loss in effectiveness which demonstrates they are over-parameterized. The contribution of this work is two-fold. The first is a method for approximating a multivariate Bernoulli random variable by means of a deterministic and differentiable transformation of any real-valued multivariate random variable. The second is a method for model selection by element-wise multiplication of parameters with approximate binary gates that may be computed deterministically or stochastically and take on exact zero values. Sparsity is encouraged by the inclusion of a surrogate regularization to the $L_0$ loss. Since the method is differentiable it enables straightforward and efficient learning of model architectures by an empirical risk minimization procedure with stochastic gradient descent and theoretically enables conditional computation during training. The method also supports any arbitrary group sparsity over parameters or activations and therefore offers a framework for unstructured or flexible structured model pruning. To conclude experiments are performed to demonstrate the effectiveness of the proposed approach.

\end{abstract}
\end{frontmatter}

\section{Introduction}
\label{s:Introduction}
Modern neural network architectures have achieved state-of-the-art results in various learning tasks and have the capacity to approximate highly complex functions. Typically such models have many millions of parameters which are densely connected. These dense structures enable efficient parallel computing using specialized software and hardware however they also require substantial resources to store the models and execute which precludes loading such models on hardware having modest resources such as low power embedded devices. Furthermore these large architectures can be pruned significantly without substantial loss in effectiveness which demonstrates that they are highly over-parameterized \cite{liu2018rethinking}. Due to being over-parameterized they are prone to overfitting which can lead to poor generalization \cite{45820}. Model compression is an approach that aims to address these aforementioned drawbacks by reducing the number of parameters while minimizing performance loss to an acceptable degree. There are many approaches suggested for model pruning or compression including weight pruning \cite{10.5555/2969239.2969366}, network quantization \cite{courbariaux2016binarized}, architecture learning \cite{DBLP:conf/bmvc/SrinivasB16}, distilling knowledge \cite{44873}, structured pruning \cite{NIPS2016_6504} and $L_0$ regularization \cite{louizos2018learning}. 

Sparse representation is related to the notion that many of the explanatory factors are irrelevant and therefore are represented by many zeros \cite{journals/corr/BengioLC13}. Consequntly a particularly useful approach is the structured or unstructured binary gating of parameters. This approach enables parameters to take on exact zero values thus enabling conditional computational savings during training without imposing additional constraints on the parameters \cite{liu2018rethinking, journals/corr/BengioLC13}. However many learning algorithms such as neural networks utilize gradient-based optimizers such as the back-propagation algorithm \citep{Rumelhart:1986we}. Models that are designed to have a continuous relationship between parameters and the training objective enable the computation of exact gradients which in turn enable efficient optimization \cite{journals/corr/BengioLC13}. The challenge is then combining discrete binary gates acting on the parameters by means of gate-parameter-wise multiplication with efficient gradient based optimization. Deterministic discreteness can be smoothed and approximated reasonably well with the softmax and sigmoid functions for categorical and binary states respectively. However if a distribution over discrete states is needed, there is no clear solution \cite{maddison2016concrete, journals/corr/BengioLC13}. Therefore, the addition of useful continuous approximations to discrete random variables can be important for the development of new models. A practical way to overcome the limitations related to including discrete states in the computation graph is by the inclusion and the optimization of stochastic nodes and taking samples of their states. A well known method is the reparameterization of stochastic nodes by deterministic functions of their parameters and a fixed noise distribution \cite{Kingma2014, pmlr-v32-rezende14}.

The main contribution of this work is two-fold. The first is a method for approximating a multivariate Bernoulli random variable by means of a deterministic and differentiable transformation of any real-valued multivariate random variable. The second is a deterministic method for model selection by element-wise multiplication of parameters with approximate binary gates that can take on exact zero values. Since the method is differentiable the gates are learned jointly with the model parameters during training by means of empirical risk minimization and theoretically offers the benefit of conditional computation during training. Note the term \emph{differential} is used in this paper in the context of training neural networks, i.e. allowing a small number of points where the first order derivatives do not exist. This is typically due to the very common use of rectifiers in the calculation graph such as the ubiquitous Relu activation \cite{NairH10}.

\section{DiffPrune}
\label{S:DiffPrune}
Let $\boldsymbol{\theta}$ be the parameters of a hypothesis $h(\cdot,\boldsymbol{\theta}): X \rightarrow Y$ such as a neural network and let $\mathcal{D}$ be a training set consisting of $N$ i.i.d. instances $\{(x_1,y_1),\ldots, (x_N,y_N)\}$. The empirical risk $\mathcal{R}$ associated with the regularized hypothesis $h(\cdot,\boldsymbol{\theta})$ is defined as:

\begin{align}
\label{eq:Symmetric C}
& \mathcal{R}_h(\boldsymbol{\theta}) = \frac{1}{N} \left(\sum_{i=1}^{N}\mathcal{L} \left( h(x_i;\boldsymbol{\theta}),y_i \right) \right) + \lambda \mathcal{L}_{reg}(\boldsymbol{\theta}) \\
& \boldsymbol{\theta}^{*} = \argmin_{\boldsymbol{\theta}} \mathcal{R}_h(\boldsymbol{\theta})
\end{align}

Where $\mathcal{L}:Y \times Y \rightarrow \mathbb{R}_{\geq 0}$ is a loss function that measures the discrepancy between the true value $y_i$ and the predicted outcome $\hat{y}_i = h(x_i;\boldsymbol{\theta})$; $\mathcal{L}_{reg}: \mathbb{R}^{\abs{\boldsymbol{\theta}}} \rightarrow \mathbb{R}_{\geq 0}$ is a regularization loss that measures the hypothesis complexity; $\lambda \in \mathbb{R}_{\geq 0}$ is a scaling factor for the regularization loss enabling trade-off between model complexity and primary task loss; and $\abs{\boldsymbol{\theta}}$ is the cardinality of $\boldsymbol{\theta}$. The goal of the optimization learning problem is to find $\boldsymbol{\theta}^{*}$ given the hypothesis $h$ and data $\mathcal{D}$ for which the empirical risk $\mathcal{R}_h(\boldsymbol{\theta})$ is minimal. The term \textit{complexity} usually corresponds to a differentiable function of the model parameters such as the $L_1$ and $L_2$ regularizers that impose soft constraints on the parameters.

Consider the hypothesis $h$ and associated empirical risk $\mathcal{R}_h$ following re-parametrization of $\boldsymbol{\theta}$ given a partition of $\boldsymbol{\theta}$ to $M$ subsets $\boldsymbol{\theta}_1 \, , \ldots  \, , \boldsymbol{\theta}_M$:

\begin{align}
& \boldsymbol{\theta} = \hat{\boldsymbol{\theta}} \odot \hat{\boldsymbol{\pi}} \, , \quad \hat{\boldsymbol{\theta}} = \bigcup\limits_{j=1}^{M} \boldsymbol{\theta}_j \, , \quad \hat{\boldsymbol{\pi}} = \bigcup\limits_{j=1}^{M} \boldsymbol{\pi}_j \, , \quad  \boldsymbol{\pi}_j \in \{0,1\}^{\abs{\boldsymbol{\theta}_j}} \, , \quad \abs{\boldsymbol{\theta}_j} \geq 2 \, , \label{eq:risk_pi_a} \\
& \mathcal{R}_{h}(\hat{\boldsymbol{\theta}}, \hat{\boldsymbol{\pi}}) = \frac{1}{N} \left(\sum_{i=1}^{N}\mathcal{L} \left( h(x_i;\cup_{j=1}^{M}\boldsymbol{\theta}_j \odot \boldsymbol{\pi}_j),y_i \right) \right) + \sum\limits_{j=1}^M \lambda_j \mathcal{L}_{reg}(\boldsymbol{\theta}_j, \boldsymbol{\pi}_j)\label{eq:risk_pi_b} \\
& \hat{\boldsymbol{\theta}}^{*} , \hat{\boldsymbol{\pi}}^{*} = \argmin_{\hat{\boldsymbol{\theta}}, \hat{\boldsymbol{\pi}}} \mathcal{R}_{h} (\hat{\boldsymbol{\theta}},\hat{\boldsymbol{\pi}}) \label{eq:risk_pi_c}
\end{align}

Where $\odot$ is the element-wise Hadamard product. Let $\theta_{jk}$ and $\pi_{jk}$ denote the $k$-th element of $\boldsymbol{\theta}_j$ and $\boldsymbol{\pi}_j$ respectively, then $\pi_{jk}$ enables a binary "gate" like function over the corresponding weight $\theta_{jk}$. Unfortunately $\mathcal{R}_{h}(\hat{\boldsymbol{\theta}}, \hat{\boldsymbol{\pi}})$ is not differentiable w.r.t. $\hat{\boldsymbol{\pi}}$ due to its binary nature. As an alternative, the next section describes a deterministic differentiable relaxation of the binary constraints over the gates that allows for the gates to take on exact zeros and enables solving a surrogate minimization problem efficiently and deterministically using common gradient based optimizers.

\subsection{Approximating a multivariate Bernoulli random variable}
To benefit from conditional computation in training it is desirable for parameters to take on exact zeros. One such way to achieve this is by utilizing multiplicative gates that can take on exact zero values. Whereas gates taking on exact zero value is crucial for efficient pruning, it is not as important for gates to be exactly one when enabled since in inference the original parameters may be substituted by the multiplication result of the gate with its corresponding parameter(s). Therefore the hard constraint of gates being binary may be relaxed and replaced with a soft constraint of being approximately one when enabled. Let $S \sim q(S|\boldsymbol{\phi})$ be a real-valued multivariate random variable with parameters $\boldsymbol{\phi}$. And let $u(\cdot)$ be a differential function from the support of $S$ to the range $(0, 1)^{\abs{\mathbf{s}}}$. Equations \eqref{eq:eff_a} - \eqref{eq:eff_e} define a deterministic and differentiable transformation $g(\cdot)$ that transforms samples taken from $S$ to follow an approximate Bernoulli distribution such that $g(S) = Z \sim \tilde{b}(Z|\boldsymbol{\phi})$.
\begin{align}
\tilde{\mathbf{z}} &= \max \left( u(\mathbf{s}) - \boldsymbol{\beta} \, , 0 \right) \, , \qquad \boldsymbol{\beta} \in (0, 1)^{\abs{\mathbf{s}}} \label{eq:eff_a} \\
\mathbf{z}^+ &= \{ \tilde{z}_k \, | \, \tilde{z}_k > 0 \, , \: k= 1,\ldots , \abs{\tilde{\mathbf{z}}} \} \label{eq:eff_b} \\
\mathbf{z}^0 &= \tilde{\mathbf{z}} - \mathbf{z}^+ \label{eq:eff_c} \\
\mathbf{z}^1 &= \left( \mathbf{z}^+ - \bar{\mathbf{z}}^+ \right) * e^{-\zeta} + 1 \, , \qquad \zeta \in \mathbb{R}_{\geq 0} \label{eq:eff_d} \\
\mathbf{z} &= \mathbf{z}^0 \cup \mathbf{z}^1 \label{eq:eff_e}
\end{align}

Where the operator $-$ in equation \eqref{eq:eff_c} is the set difference and $\bar{\mathbf{z}}^+$ is the sample mean of $\mathbf{z}^+$. The transformation defined by $g(\cdot)$ conceptually comprises $\mathbf{z}$ of two partitions: $\mathbf{z}^0$ and $\mathbf{z}^1$, such that by definition all $\mathbf{z}^0 = 0$ and assuming that $\abs{\mathbf{z}^1} > 0$ then $\E ( \mathbf{z}^1 ) = 1$. The variance of $\mathbf{z}^1$ is controlled by $\zeta$ and since $ \abs{z_k^+ - z_l^+} < 1 $ it may be set as small as practically useful and can therefore approximate a binary random variable to any finite practical precision. In addition note that $Z$ becomes exactly binary in the limit as $\zeta \rightarrow \infty$.  It is clear that the distribution $\tilde{b}(Z|\boldsymbol{\phi})$ depends on the choice of $q(S|\boldsymbol{\phi}), u(\cdot)$ and $\boldsymbol{\beta}$. It is also notable that $\tilde{b}(Z|\boldsymbol{\phi})$ is strongly bimodal in the non-degenerate case even if the distribution $q(S|\boldsymbol{\phi})$ is unimodal. Finally the gradient of $\mathbf{z}$ w.r.t. $\mathbf{s}$ is non-degenerate provided that $\abs{\mathbf{z}^1} \geq 2$ i.e. there are at least two members in $\mathbf{z}^+$. \newline

There are many possible choices for $u(\cdot)$ however in this paper the focus is on two variants. The first is the element-wise $sigmoid(\cdot)$ function and the second is the $softmax(\cdot)$. \newline

\textbf{Sigmoid}: Let $u \coloneqq sigmoid(\cdot)$, the marginal probability of $z_k$ being approximately one is:

\begin{align}
p \left(z_k > 0 \right) & = p \left(\frac{1}{1+e^{-s_k}} > \beta_k \right) = p \left( s_k > - \ln {\left( 1 / \beta_k - 1 \right) } \right) \label{eq:p_enabled_z_expect_a_sigmoid} \\
& = 1 - CDF_{s_k} \left( - \ln {\left( 1 / \beta_k - 1 \right) } \right) \label{eq:p_enabled_z_expect_b_sigmoid}
\end{align}

Where $CDF_{s_k}$ is the marginal cumulative density function of $s_k$. Therefore in the case that $S$ is independent $u \coloneqq sigmoid(\cdot)$ results in the events $z_k \approx 1$ being independent, i.e. $p \left( z_k \approx 1, z_l \approx 1 \right) = p \left( z_k \approx 1 \right) p \left( z_l \approx 1 \right)$. \newline

\textbf{Softmax}: Let $u \coloneqq softmax(\cdot)$, the conditional probability of $z_k$ being approximately one given $\{S_l=s_l \, | \, l \neq k\}$ is:

\begin{align}
p \left(z_k > 0 \: | \:  \{S_l=s_l \, | \, l \neq k\} \right) = p \left( e^{s_k} / \sum_{l = 1}^{\abs{S}} e^{s_l} > \beta_k  \: | \:  \{S_l=s_l \, | \, l \neq k\} \right) \label{eq:p_enabled_z_softmax}
\end{align}

Rearranging the inequality in equation \eqref{eq:p_enabled_z_softmax}:

\begin{align}
& e^{s_k} / \sum_{l = 1}^{\abs{S}} e^{s_l} > \beta_k \\ 
& e^{s_k} > \beta_k \sum_{l = 1}^{\abs{S}} e^{s_l} \\
& \left( 1 - \beta_k \right) e^{s_k} > \beta_k \sum_{l = 1, l \neq k}^{\abs{S}} e^{s_l}  \\
& s_k > \ln \left( \frac{\beta_k}{1 - \beta_k} \sum_{l = 1, l \neq k}^{\abs{S}} e^{s_l} \right) \label{eq:p_enabled_z_softmax_ineq}
\end{align}

Therefore by combining equations \eqref{eq:p_enabled_z_softmax} and \eqref{eq:p_enabled_z_softmax_ineq} the conditional probability of $z_k$ being approximately one given $\{S_l=s_l \, | \, l \neq k\}$ is:

\begin{align}
p \left(z_k > 0 \: | \:  \{S_l=s_l \, | \, l \neq k\} \right) & = p \left( s_k > \ln \left( \frac{\beta_k}{1 - \beta_k} \sum_{l = 1, l \neq k}^{\abs{S}} e^{s_l} \right) \right) \\
& = 1 - CDF_{s_k} \left( \ln \left( \frac{\beta_k}{1 - \beta_k} \sum_{l = 1, l \neq k}^{\abs{S}} e^{s_l} \right) \right)
\end{align}

Note that the choice of $u \coloneqq softmax ( \cdot) $ leads to the events $z_k \approx 1$ having inherent dependencies regardless of the independence of $S$. \newline

The formulation in equations \eqref{eq:eff_a} - \eqref{eq:eff_e} offers a number of benefits in the context of pruning which are discussed in subsequent sections. Most importantly the empirical risk in equation \eqref{eq:risk_pi_b} can now be smoothed by replacing the binary Bernoulli gates $\hat{\boldsymbol{\pi}}$ with the approximately binary gates $\hat{\mathbf{z}}$ and thus enable the use of gradient based optimizers:

\begin{align}
& \boldsymbol{\theta} = \hat{\boldsymbol{\theta}} \odot \hat{\mathbf{z}} \, , \quad \hat{\boldsymbol{\theta}} = \bigcup\limits_{j=1}^{M} \boldsymbol{\theta}_j \, , \quad \hat{\mathbf{z}} = \bigcup\limits_{j=1}^{M} \mathbf{z}_j \, , \quad  \mathbf{z}_j \in \{0,\tilde{1}\}^{\abs{\boldsymbol{\theta}_j}} \, , \quad \abs{\boldsymbol{\theta}_j} \geq 2  \, , \quad Z_j \sim \tilde{b}(g(S_j)\, | \, \boldsymbol{\phi}_j) \label{eq:risk_z_a} \\
& \mathcal{R}_{h}(\hat{\boldsymbol{\theta}}, \hat{\boldsymbol{\phi}}) = \frac{1}{N} \left(\sum_{i=1}^{N}\mathcal{L} \left( h(x_i;\cup_{j=1}^{M}\boldsymbol{\theta}_j \odot \mathbf{z}_j),y_i \right) \right) + \sum\limits_{j=1}^M \lambda_j \mathcal{L}_{reg}(\boldsymbol{\theta}_j,\mathbf{z}_j)\label{eq:risk_z_b} \\
& \hat{\boldsymbol{\theta}}^{*} , \hat{\boldsymbol{\phi}}^{*} = \argmin_{\hat{\boldsymbol{\theta}}, \hat{\boldsymbol{\phi}}} \mathcal{R}_{h} (\hat{\boldsymbol{\theta}},\hat{\boldsymbol{\phi}}) \label{eq:risk_z_c}
\end{align}

\subsection{Gates}
\label{S:Gates}
By transforming the samples of continuous distributions that allow for the reparameterization trick \cite{Kingma2014, pmlr-v32-rezende14} it is possible to express the stochastic objective in \eqref{eq:risk_z_b} as an expectation over a deterministic differentiable transformation $f(\cdot)$ of the parameters $\hat{\boldsymbol{\phi}}$ and a parameter free noise distribution $p ( \boldsymbol{\epsilon} )$ and perform Monte Carlo approximation to the intractable expectation over the noise distribution. Whereas injecting noise to the gradients estimation can be useful such as in the case of dropout \cite{10.5555/2627435.2670313}, it may also have the undesirable effect of increasing the variance of the estimator \cite{Huang_2020_CVPR_Workshops}. Furthermore estimation by means of sampling to produce expectations has the drawback of increasing the computational requirements. Consequently it is proposed to choose $q(S|\boldsymbol{\phi})$ such that no sampling is required and the objective \eqref{eq:risk_z_b} becomes deterministic. Let $S_j$ follow an isotropic normal distribution $\mathcal{N}(\boldsymbol{\mu}_j, \sigma^2 I)$, it is thus possible to perform a deterministic and differentiable maximum likelihood estimation of the gates values simply by computing $\mathbf{z}_j = g(\boldsymbol{\phi}_j) = g(\boldsymbol{\mu}_j)$. Note that $\boldsymbol{\mu}_j$ are nuisance parameters that are only required during training and may be pruned with the corresponding model parameter(s).

It is perhaps useful to provide an intuitive interpretation of $g(\cdot)$ in the context of computing gates over the model parameters. One way of many to select which gates are enabled is to rely on the notion of relative "utility" of the parameters in $\boldsymbol{\theta}_j$ and to rank them based on the probability that a parameter $\theta_{jk}$ is the most "useful" or "important" in the set. In the case where $u(\cdot) \coloneqq softmax(\cdot)$ the random variable $S_j$ is then interpreted as the logits of a categorical random variable denoting which of the elements of $\boldsymbol{\theta}_j$ has the "most utility". In the case where $u(\cdot) \coloneqq sigmoid(\cdot)$ the random variable $S_j$ is interpreted as the logits of a multivariate Bernoulli random variable denoting the importance probability for each of the weights independently. In this context, for simplicity, $\boldsymbol{\beta}_j$ can be reduced to a scaler and act as a hyperparameter specifying the probability selection threshold.

Given the choice of $S \sim \mathcal{N}(\boldsymbol{\mu}, \sigma^2 I)$ and due to using deterministic maximum likelihood sampling it follows that:

\begin{align}
p \left(z_k > 0 \right) = 1 - CDF_{\mathcal{N}(\mu_k, \sigma^2)} \left( - \ln {\left( 1 / \beta - 1 \right) } \right)
\end{align}
and:
\begin{align}
p \left( z_k > 0 \right) & = 1 - CDF_{\mathcal{N}(\mu_k, \sigma^2)} \left( \ln \left( \frac{\beta}{1 - \beta} \sum_{l = 1, l \neq k}^{\abs{\boldsymbol{\mu}}} e^{\mu_l} \right) \right)
\end{align}

for when $u(\cdot) \coloneqq sigmoid(\cdot)$ and $u(\cdot) \coloneqq softmax(\cdot)$ respectively. Note the edge case resulting in degenerate gradients when there is at most one active gate in a partition i.e. $\abs{\mathbf{z}^1} \leq 1$. However in practice the need to prune all parameters but one is unlikely and therefore this limitation is not expected to cause issues in application of this method.

\subsection{Regularization}
If no regularization is performed, i.e. only the the first term in equation \eqref{eq:risk_z_b} is present, the gates will converge to some minima in respect to the primary objective with no additional restrictions. This is useful for configuration free model selection that is likely to retain the cumbersome model's accuracy. However in many cases it is desirable to encourage the model to achieve a higher sparsity with the trade-off of lower accuracy by the introduction of regularization loss. Note that for simplifying the notation in the remainder of this section it is assumed that the parameters are not partitioned and therefore the partition subscript is omitted. \newline

\textbf{$\mathbf{L_0}$ regularizer}: The $L_0$ loss is simply the number of parameters of the model that are different from zero. This regularisation loss has desirable attributes in the context of pruning since it "punishes" the model for having non-zero parameters with no further restrictions and does not induce shrinkage. Assuming that $p(w_k = 0) = 0$ and due to $\E (\mathbf{z}^1 ) = 1$ it is possible to compute the $L_0$ loss simply as $L_0(\mathbf{z}) = \sum \mathbf{z}$. Unfortunately this will yield zero gradients w.r.t. $\boldsymbol{\mu}$, or w.r.t. each $\boldsymbol{\mu}_j$ in the case where the parameters are partitioned. Instead the approach suggested in \cite{louizos2018learning} is adopted and rather than directly computing the number of enabled gates the expected number of enabled gates, whether sampling or not, is used as a surrogate loss. The expected $L_0$ loss over the parameters is obtained by summing the marginal probability or marginal conditional probability in the case of $u(\cdot) \coloneqq softmax(\cdot)$ of every gate being enabled:

\begin{align}
& \mathcal{L}_{reg}(\mathbf{z}) = \sum_{k = 1}^{\abs{Z}} p \left(z_k > 0 \right)
\end{align}

Or in the case where the parameters of the model are partitioned:

\begin{align}
& \mathcal{L}_{reg}(\mathbf{z}) = \sum_{j = 1}^{M} \sum_{k = 1}^{\abs{Z_{j}}} p \left(z_{jk} > 0 \right) \label{eq:l0_loss_total_all_partitions}
\end{align}

\textbf{Gaussian dropout}: Experimental results discussed in subsequent sections demonstrate that applying Gaussian dropout \cite{10.5555/2627435.2670313} to $\boldsymbol{\mu}$ has a substantial effect on the results, both the compression rate and the overall accuracy of the model. From a practical perspective injecting noise has the side effect of mild stochastic binary dropout due to gates being randomly enabled and disabled. From a theoretical point of view the product of the two random variables, $\mathcal{N}(\boldsymbol{\mu}, \sigma^2 I)$ and $\mathcal{N}(1, \sqrt{p/(1-p)} )$ does not follow a multivariate normal distribution however since in practice the ML sampling yields a constant $\boldsymbol{\mu}$ the resulting noise follows an uncorrelated multivariate normal distribution where the standard deviation of each of the components is proportional to the absolute value of its mean.

\subsection{Group sparsity}
In many applications it is desired to enforce group pruning of parameters since this can enable more substantial computational savings in practice with existing hardware and software libraries \cite{NIPS2016_6504}. A common scenario is the pruning of the activations computed by a layer, e.g. neuron pruning. Other applications require jointly pruning groups of activations such as when performing filter pruning of a CNN \cite{DBLP:conf/iclr/0022KDSG17}. DiffPrune can be used to this extent by virtue of associating a gate in $\mathbf{z}$ with a group of parameters or activations. Note the cardinality of nuisance parameters $\abs{\boldsymbol{\mu}}$ is equal to the number of gates which is typically substantially less than required for unstructured parameter pruning.

\subsection{Initialization and configuration}
\textbf{Partitioning and sparsity}: The proposed method enables arbitrary pruning schemes in respect to any combination of parameters, groups of parameters, activations and groups of activations. Furthermore by including the regularization loss in the objective different target compression rates for each partition can be specified by configuration of the parameters $\boldsymbol{\lambda}$. Larger values of $\lambda_j$ will accelerate pruning by balancing the magnitude of the $L_0$ regularization terms against the model's primary loss. Both the partition scheme and the values of the hyperparameters are dependent on the objectives of the practitioner, the data and the architecture. \newline

\textbf{Initializing $\boldsymbol{\mu}_j$}: Let $\boldsymbol{\mu}_j^{(0)}$ denote the initial value of $\boldsymbol{\mu}_j$. $\boldsymbol{\mu}_j^{(0)}$ can be considered as defining a priori probability distribution on the "utility" of weights in $\boldsymbol{\theta}_j$ and for certain values can result in pruning of parameters pre-training. Typically through application of the principle of indifference a uniform distribution is the default choice in cases such as this unless there is a particular reason to do otherwise. However for effective backpropagation it is required that the parameters $\boldsymbol{\mu}_j$ are initialized randomly and have both positive and negative values. Therefore it is recommended for $\boldsymbol{\mu}_j^{(0)}$ to have a mean of zero and small variance so that it is close to being uniform and works well in the context of the SGD optimization framework. In experiments done as part of this work a truncated normal distribution with a standard deviation of $0.05$ is used. \newline

\textbf{Selecting $\beta_j$}: Effectively acting as a pruning selection threshold where parameters that have a utility probability not greater than $\beta_j$ are pruned. Let $\mathbf{p}_j^{(0)} = u(\boldsymbol{\mu}_j^{(0)})$. Given no a priori preference as to which parameters are to be pruned it is desirable to select $\beta_j$ such that $p_{jk} - \beta_j > 0 \, , \: \forall p_{jk} \in \mathbf{p}_j^{(0)}$ as this leads to no parameters being pruned pre-training. Therefore $\beta_j$ can be selected heuristically during the initialization stage without the need to configure it by the practitioner. In theory it can be said that lower values of $\beta_j$ provide the model the opportunity to train for longer before pruning commences. In experiments done in this work values of $\beta_j = 0.99 \min ( \mathbf{p}_j^{(0)} ) $ are used. \newline

\textbf{Learning $\zeta_j$}: Since $ \abs{z_{jk}^+ - z_{jl}^+} < 1 $ it is possible in theory to reduce the variance of the enabled gates $\mathbf{z}_j^1$ to an arbitrary value by increasing $\zeta_j$. In this work however $\zeta_j$ is learned together with the other parameters of the model. \newline

\textbf{Dropout rate}: The Gaussian dropout rate is reparameterized as $\sqrt{\sigma(\eta_j)/(1-\sigma(\eta_j))}$ where $\eta_j \in \mathbb{R}$ can be learned jointly with the other parameters of the model and $\sigma(\cdot)$ is the sigmoid function. To enable backpropagation for training $\eta_j$ through the sampling stage gradient estimates are obtained by a single sample Monte Carlo and the reparameterization trick \cite{Kingma2014, pmlr-v32-rezende14}.

\section{Related work}
\label{S:RelatedWork}
Methods for approximating binary discrete random variables that can represent exact zero by rectifications of continuous random variables have been suggested before in various contexts. These include rectifying samples taken from the logistic distribution by a Relu for conditional computation \cite{journals/corr/BengioLC13} however the resulting distribution is not strongly bi-modal and therefore cannot represent effectively binary random variables. The hard concrete distribution proposed in \cite{louizos2018learning} can emulate binary variables more closely by hard clipping a stretched concrete random variable \cite{maddison2016concrete, 45822}. This approach can represent both exact zero and one however has some drawbacks, the gradients suffer from high-variance estimates; it is limited only to its own Concrete distribution and does not generalize to other multivariate distributions \cite{Huang_2020_CVPR_Workshops}. Furthermore it is continuous in $[0,1]$ and the hard rectifiers on both extremes substantially increase the sparsity of gradients during backpropagation. Expanding on this framework is the work of \cite{Huang_2020_CVPR_Workshops} where any mixture pair of distributions converging to $\delta(0)$ and $\delta(1)$ can be used to construct smoothed binary gates. By contrast the method proposed here can transform any underlying distribution including unimodal distributions into practically discrete binary random variables and maintain the tractability of continuous optimization. Furthermore due to the use of the $softmax(\cdot)$ function that creates dependency between the components of the emulated binary random variable even components that are zero are affected by gradient updates and do not suffer to the same extent from gradient "dead zone". A different approach is proposed in \cite{rolfe2017discrete} for gradient based optimization of discrete random variables by smoothing transformations of binary latent variables with continuous noise in a way that allows for gradients reparameterization.

Many methods have been proposed for compression and sparsification of neural networks during the last three decades. Due to the large volume of work it is not possible to mention the absolute majority of methods and instead the reader is referred to \cite{journals/corr/abs-1710-09282, DBLP:conf/mlsys/BlalockOFG20} for a comprehensive review. Most methods proposed in the literature follow a three stage procedure where (i) an over-parameterized large network is trained to partial or full convergence; (ii) following the training redundant weights or neurons are pruned based on heuristics; (iii) the network is trained again possibly from scratch to "fine-tune" the model \cite{liu2018rethinking,10.5555/2969239.2969366, DBLP:conf/iclr/MolchanovTKAK17, DBLP:conf/iccv/LuoWL17}. This procedure may repeat a number of times to iteratively increase the model compression. One major shortcoming of many of the methods that follow this three stage procedure is they require training the over specified network to convergence which precludes conditional computational savings during training \cite{liu2018rethinking, journals/corr/BengioLC13}.

Similar to the method here in the context of network pruning is the $L_0$ regularization for neural networks proposed in \cite{louizos2018learning, Huang_2020_CVPR_Workshops} that prunes the network during training by the introduction of smooth Bernoulli stochastic gates. These frameworks however resort to Monte Carlo sampling to estimate the posterior probability distribution over the parameters. Other related methods are proposed in \cite{srinivas2016training, srinivas2016generalized}, however the main difference and drawback of these methods is that they are non-differential and use the high variance straight-through estimator \cite{journals/corr/BengioLC13} to compute the gradients of the gate distribution parameters. A rather differnet method that performs online pruning while training is proposed in \cite{journals/corr/abs-1710-01878} where magnitude based heuristics determine which weights to prune.

\section{Experiments}
It is generaly not feasible to perform exact comparisons between published methods due to the discrepancies related to differing frameworks, partial information, pre-processing pipelines and stochastic aspects \cite{DBLP:conf/mlsys/BlalockOFG20}. However to demonstrate the effectiveness of the proposed method an attempt is made to compare it with the results published in \cite{louizos2018learning, Huang_2020_CVPR_Workshops} using similar architectures, datasets and preprocessing. The models are implemented in TensorFlow \cite{tensorflow2015-whitepaper} using standard Dense and Conv2D layers with additional custom DiffPrune layers added between the standard layers to perform neuron and group pruning on the activations of the standard layers. Conditional computation is not performed in practice however an indicative number of FLOPs per training step is recorded and used to demonstrate potential theoretical benefits. All image datasets were taken from TensorFlow Datasets \cite{TFDS} with the default train/test split. The source code is publicly available at \href{https://bitbucket.org/YanivShu/diffprune_public}{https://bitbucket.org/YanivShu/diffprune\_public}. Note there was no attempt to perform an exhaustive search of hyperparameters for the best possible result therefore these results should be taken as indicative only.

\subsection{LeNet5 MNIST classification}
The first experiment is the toy classification task of MNIST using the basic CNN LeNet5 \cite{Lecun98gradient-basedlearning} with an original architecture of 20-50-800-500. The results for two variants of DiffPrune are reported in table \ref{tab:lenet5}. The first has no regularization loss and $\boldsymbol{\eta}$ is trained with $\eta^{(0)}=0$ for all partitions with $u \coloneqq softmax ( \cdot )$ and thus demonstrates the capabilities of the method to perform configuration free "no fuss" model selection during training. The second has a regularization loss term added to the objective as per equation \eqref{eq:l0_loss_total_all_partitions} with $u \coloneqq sigmoid ( \cdot )$, $\boldsymbol{\lambda} = \{ 10^{-5}, 10^{-5}, 2*10^{-5}, 2*10^{-5} \}$ and $\boldsymbol{\eta}$ was trained with $\eta^{(0)} \approx -1.734$ for all partitions. Training was performed for 200 epochs with testing following the completion of each epoch, of which a representative result is reported. The batch size was 100 and optimization is performed using Adam with default parameters and a learning rate of $0.0005$.

\begin{table}[t]
\centering
\begin{tabular}{l c c c c } 
\toprule
Method & Pruned Arch. & Error \% \\ [0.5ex] 
\midrule
Sparse VD \cite{pmlr-v70-molchanov17a} & 14-19-242-131 & 1.0 \\
GL \cite{NIPS2016_6504} & 3-12-192-500 & 1.0 \\
GD \cite{srinivas2016generalized} & 7-13-208-16 & 1.1 \\
SBP \cite{NIPS2017_7254} & 3-18-284-283 & 0.9 \\
BC-GNJ \cite{NIPS2017_6921} & 8-13-88-13 & 1.0 \\
BC-GHS \cite{NIPS2017_6921} & 5-10-76-16 & 1.0 \\
$L_0$ \cite{louizos2018learning} & 20-25-45-462 & 0.9 \\
$L_{0, \, \lambda \text{sep}}$ \cite{louizos2018learning} & 9-18-65-25 & 1.0 \\
MDR-ExpUni \cite{Huang_2020_CVPR_Workshops} & 6-8-72-31 & 1.0 \\
\midrule
DiffPrune & 11-30-496-309 & 0.6 \\ 
DiffPrune\textsubscript{reg} & 10-20-71-35 & 1.0 \\
\bottomrule
\end{tabular}
\caption{\label{tab:lenet5}Comparison of architecture learned and error rates by methods compared in \cite{louizos2018learning, Huang_2020_CVPR_Workshops} and DiffPrune on the MNIST classification task using LeNet5. The DiffPrune\textsubscript{reg} variant has a regularization loss term incorporated to the objective.} 
\end{table}

\subsection{Wide Residual Network CIFAR10 classification}

The second experiment is the classification task of the CIFAR10 dataset using a pruned Wide Residual Network \cite{BMVC2016_87}. The baseline architecture is the WRN-28-10 with no dropout and an attempt is made to maintain as closely as possible the training procedue and settings described in \cite{BMVC2016_87} with some minor differences to the data augmentation pipeline and a smaller batch size of 120. Filter pruning is applied in all groups excluding group conv1, and in all blocks on the hidden Conv2D layer as well as jointly on the outputs of the second Conv2D layer and the skip connection convolution layer. Applying pruning jointly to both convolution layers at the same activation plane results in the same activation dimensions and enables conditional computation benefits. In practice this is implemented by adding a DiffPrune layer to prune the addition results of the convolution layers activations. The DiffPrune layers are all configured with $u \coloneqq sigmoid ( \cdot )$ and to enable a fair comparison no Gaussian or other dropout was applied. 

Two results are reported varying in their regularization settings. The first result demonstrates pruning with no accuracy loss and the second demonstrates more aggressive regularization with little loss of accuracy. These two variants are referred to as "low compression" and "high compression" respectively.

For the low compression variant the error rate achieved on the test set was the same as reported in \cite{BMVC2016_87} of $\approx0.42\%$. The DiffPrune layers in all blocks are configured with $\boldsymbol{\lambda} = \{ 10^{-5}, 10^{-5}, 5 * 10^{-5} \}$ for each of the groups conv2, conv3 and conv4 respectively. The pruned architecture learned is: 16-[(133,122)-(148,105)-(137,101)-(106,58)]-[(301,255)-(316,248)-(310,240)-(318,214)]-[(618,543)-(624,578)-(620,600)-(615,635)].

The high compression variant achieves an error rate of $\approx4.5\%$. The DiffPrune layers in all blocks are configured with $\boldsymbol{\lambda} = \{ 2 * 10^{-5}, 7 * 10^{-5}, 3 * 10^{-4} \}$ for each of the groups conv2, conv3 and conv4 respectively. The pruned architecture learned during training has substantially less enabled filters in deeper layers of the network: 16-[(127,120)-(137,112)-(139,102)-(113,61)]-[(286,230)-(296,235)-(289,238)-(304,231)]-[(148,180)-(162,165)-(167,143)-(153,169)].

\begin{figure}[h!]
\centering
\begin{subfigure}{.5\textwidth}
    \centering
    \includegraphics[width=1.0\textwidth]{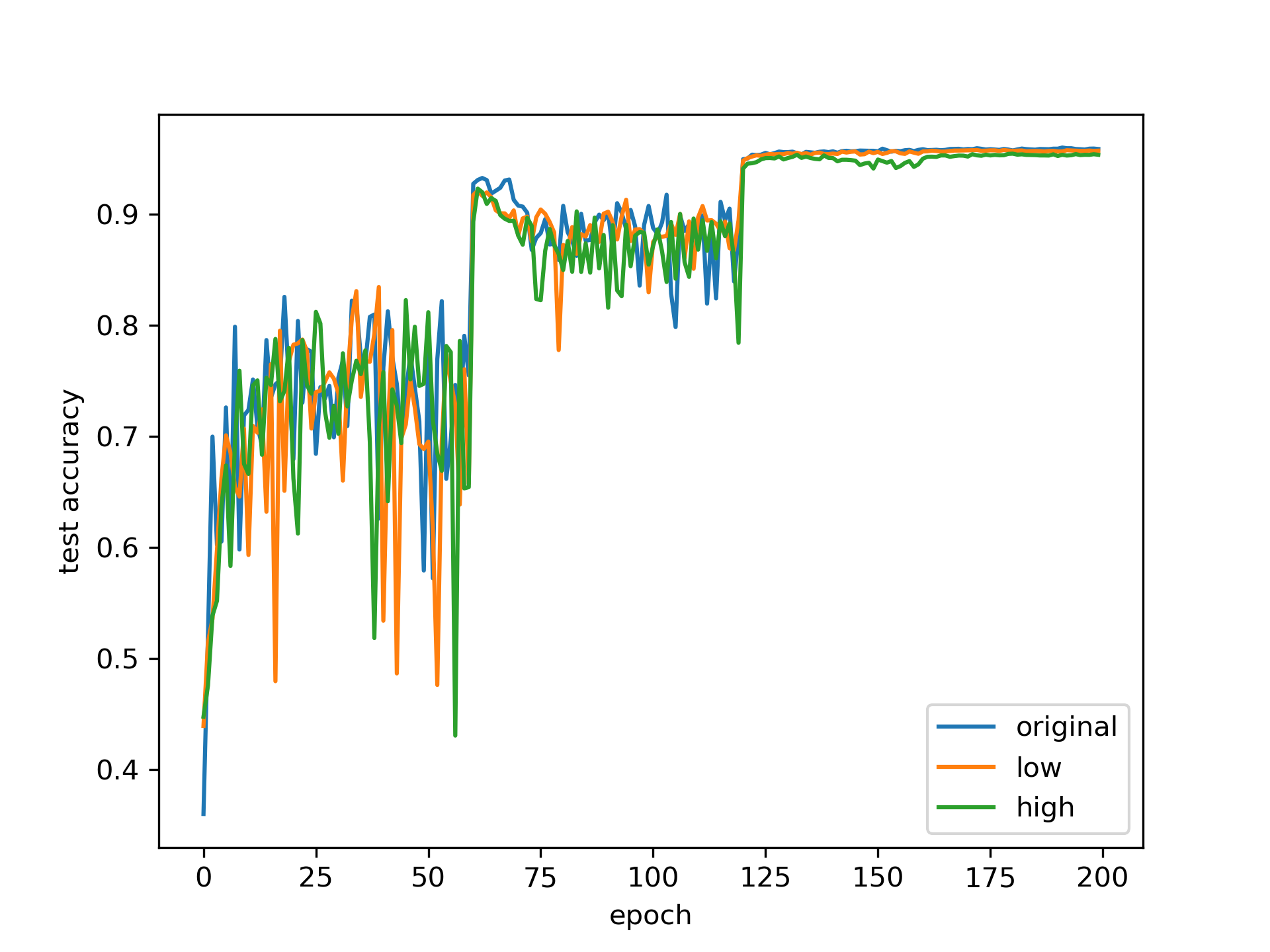}
    \caption{}
\end{subfigure}%
\begin{subfigure}{.5\textwidth}
    \centering
    \includegraphics[width=1.0\textwidth]{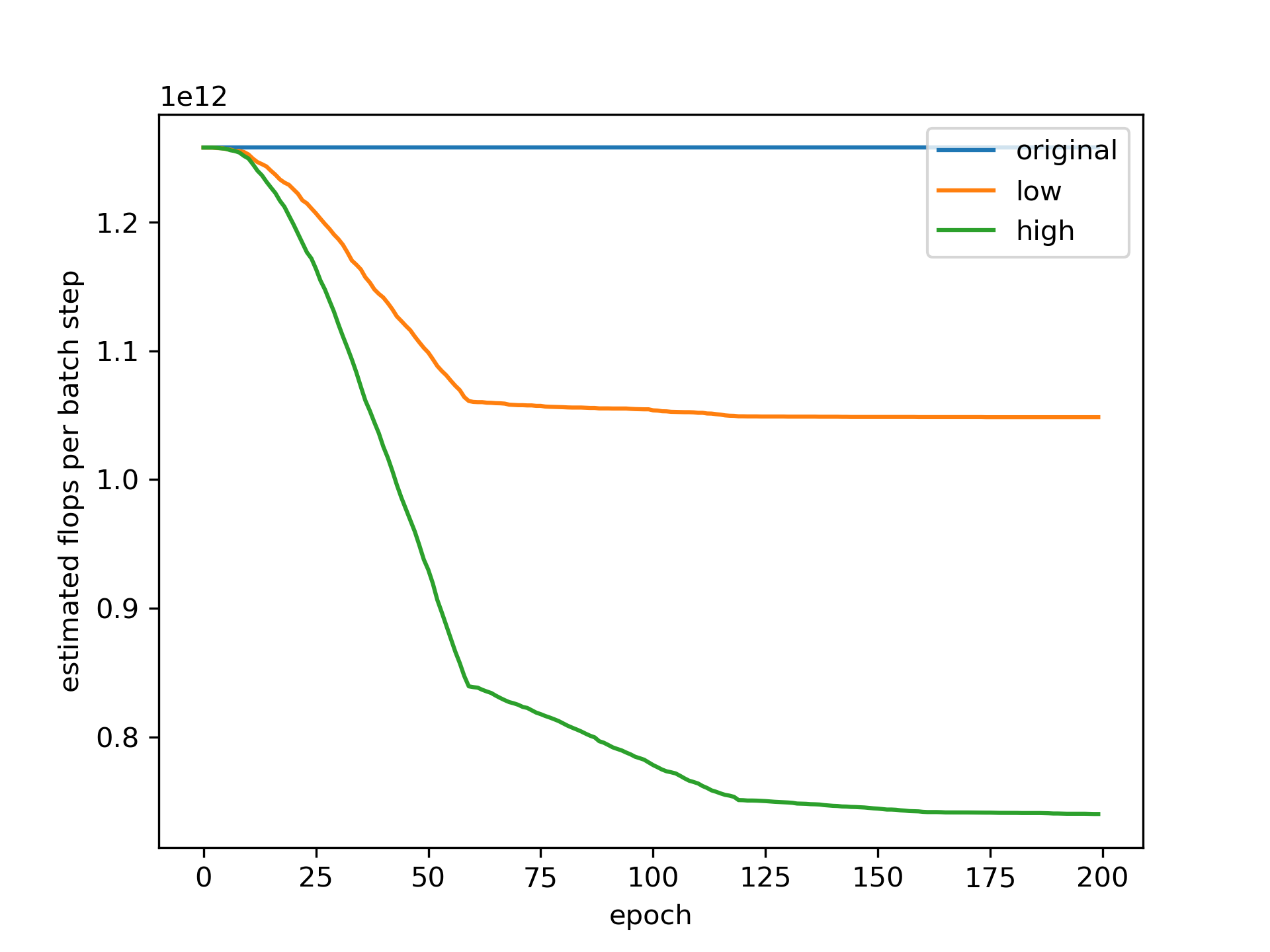}
    \caption{}
\end{subfigure}
\caption[short]{(a) Accuracy measured on the test set for the CIFAR10 classification task with the original and pruned WRN-28-10 wide residual network. (b) The number of estimated FLOPs for the forward pass of a mini-batch .}
\label{fig:wrn_plots}
\end{figure}

\section{Conclusion}
This paper proposes a novel distribution that approximates efficiently and effectively the multivariate Bernoulli distribution by employing a deterministic transformation to random variables with no restrictions on the underlying distribution. By choosing a continuous underlying distribution that allows for efficient ML estimation or the reparameterization trick it better emulates the binary nature of Bernoulli distributions while enabling the use of efficient gradient based optimization in deterministic or stochastic computation graphs respectively. Furthermore a deterministic framework for model selection with approximate binary gates and exact gradients was proposed which can prune models with or without regularization whilst maintaining the error rate of the original cumbersome model. Furthermore it is demonstrated that this method allows for deterministic optimization of the $L_0$ norm of parametric models by smoothing the combinatorial problem with the proposed approximate Bernoulli distribution. Experiments demonstrate that the proposed method can competitively sparsify neural networks in comparison to current approaches while theoretically allowing for conditional computation in training.

\bibliographystyle{abbrv}

\bibliography{refs}

\end{document}